\documentclass{article}

\usepackage{arxiv}

\usepackage[utf8]{inputenc} 
\usepackage[T1]{fontenc}    
\usepackage{hyperref}       
\usepackage{url}            
\usepackage{booktabs}       
\usepackage{amsfonts}       
\usepackage{nicefrac}       
\usepackage{microtype}      
\usepackage{graphicx}
\usepackage[numbers]{natbib}
\usepackage{xcolor}
\usepackage{amsmath}
\usepackage{tikz}
\usepackage{xcolor}
\usetikzlibrary{positioning, shapes, arrows, fit, backgrounds, calc, decorations.pathreplacing, calligraphy}

\title{MONET - Virtual Cell Painting of Brightfield Images and Time Lapses Using Reference Consistent Diffusion}

\author{
  Alexander Peysakhovich\textsuperscript{1,2} \\
  \And
  William Berman\textsuperscript{1,2} \\
  \And
  Joseph Rufo\textsuperscript{3} \\
  \And
  Felix Wong\textsuperscript{1} \\
  \And
  Maxwell Z. Wilson\textsuperscript{1,3} \\
}

\renewcommand{\shorttitle}{MONET - Virtual Cell Painting}

\begin{document}
\maketitle
\renewcommand{\thefootnote}{}
\footnotetext{\begin{minipage}{\textwidth}
\textsuperscript{1}Integrated Biosciences, Redwood City, CA \\
\textsuperscript{2}Sutter Hill Ventures, Palo Alto, CA \\
\textsuperscript{3}University of California - Santa Barbara, Santa Barbara, CA
\end{minipage}}
\renewcommand{\thefootnote}{\arabic{footnote}}

\begin{abstract}
Cell painting is a popular technique for creating human-interpretable, high-contrast images of cell morphology. There are two major issues with cell paint: (1) it is labor-intensive and (2) it requires chemical fixation, making the study of cell dynamics impossible. We train a diffusion model (Morphological Observation Neural Enhancement Tool, or MONET) on a large dataset to predict cell paint channels from brightfield images. We show that model quality improves with scale. The model uses a consistency architecture to generate time-lapse videos, despite the impossibility of obtaining cell paint video training data. In addition, we show that this architecture enables a form of in-context learning, allowing the model to partially transfer to out-of-distribution cell lines and imaging protocols. Virtual cell painting is not intended to replace physical cell painting completely, but to act as a complementary tool enabling novel workflows in biological research.
\end{abstract}

\section{Introduction}
Cell painting is a common technique used to determine and catalog the effects of perturbations on cellular biology. Cells are first incubated with a mitochondrial dye in live culture, then chemically fixed, after which additional dyes that label specific subcellular structures are applied. Cell painting creates high contrast images which can then be used to track the state and health of cellular organelles and other cellular subsystems by analyzing their morphology under various conditions and perturbations \citep{bray2016cell, cimini2023optimizing, krentzel2023deep, seal2025cell}.

Prior work in machine learning shows that much of the morphological information that is highlighted in cell paint is actually available in simple brightfield imaging \citep{harrison2023evaluating,xing2024can}. However, cell painting remains popular because the composite fluorescent images are human-interpretable. They highlight nuclei, cytoskeleton, mitochondria, and other organelles in a way that lets researchers visually recognize phenotypes, spot artifacts, and rapidly triage interesting perturbations.

Cell painting has two main drawbacks. It is labor intensive, involving live-cell mitochondrial staining and sequential application of two dye mixes with multiple incubations and washes, and it requires chemical fixation, which precludes studying cell dynamics or time-lapse behavior.

This leads to a practical question: can we computationally generate cell paint directly from brightfield? Seminal work has shown at that it is indeed possible to generate cell paint images either from brightfield or from conditioning on perturbations\citep{christiansen2018silico,ounkomol2018label,wieslander2021learning,cross2022label,cross2023class,navidi2024morphodiff,xing2025artificial}. We scale this idea by increasing model size (from 31 million parameters in \citep{cross2022label} to 250 million), increasing training data size (from $3$ million images in \citep{xing2025artificial} to over $8$ million images), increasing resolution (from $256$ squared pixels in \citep{cross2022label,xing2025artificial}  to $512$ squared), and using a modern setup (diffusion with flow matching \citep{lipman2022flow}) and perform virtual cell painting directly from brightfield. 

In addition to scaling the model and data, we focus on the problem of virtually cell painting brightfield timelapses. Here, no real dataset of paired (brightfield, cellpaint) timelapses can exist. Unfortunately, generating each frame independently leads to videos with many artifacts. For this reason, we introduce an architecture called `reference consistency' where the generation is optionally conditioned on a different ground truth (brightfield, cellpaint) example. At training time, we use a different view / augmentation of the same example as the conditioning. At inference time, we generate the first frame of a timelapse without conditioning and then condition subsequent frames on the first frame. 

We perform visual evaluations of the images and timelapses it generates - the reader can see many examples from our model at \url{https://www.thiscellpaintingdoesnotexist.com}. All of these examples come from datasets that has not been seen by the model during training. To generate time lapses, we construct a dataset by brightfield imaging growing U2OS cells every $10$ minutes for $24$ hours.

Arbitrary choices (e.g. length of dye exposure, illumination) by the experimenter will affect resulting cell paint images. Thus, there is no single `right' cell paint for a given brightfield. For this reason, perceptual metrics of generated images do not tell the whole story.

To deal with this issue, we consider an evaluation that looks at how important information is preserved in virtual vs generated brightfields. To evaluate the image-only model, we use a dataset of cells treated with known compounds. Each perturbation has a labeled mechanism of action (MOA). We compare the accuracy of a classifier trained to predict MOA from image using the ground truth cell paint images or the virtual cell paint. 

To evaluate the usefulness of our consistency module, we compare the frame-to-frame consistency of time lapses generated with and without the use of our reference consistency, showing that indeed more visual artifacts are introduced by naively generating one frame at a time.

We consider whether there is room for further scaling the model. We train MONET models of sizes ranging from $30$ million to $250$ million parameters on the same dataset. Our qualitative and quantitative evaluations show there appears to be room for further scaling in both model and data.

Finally, we consider the problem of domain adaptation by looking at an in-house generated cell paint dataset of a cell line (human fibroblasts) that our model has never seen during training imaged on hardware that is very different from the pre-training data. We show that unlocking the full performance of the model requires domain specific fine tuning but that our reference consistency architecture gives the model some amount of ability to adapt to the new domain via a form of in-context learning. 

We do not claim that virtual cell painting can completely replace cell painting in all situations. However, it is a useful tool that enables new workflows, for example, imaging human interpretable time-lapses in brightfield or automated high throughput screening with brightfield and the use of virtual cell painting for human oversight and spot-checking of results. 

\section{Model and Data}
\begin{figure}[h!]
    \centering
    \includegraphics[width=0.7\linewidth]{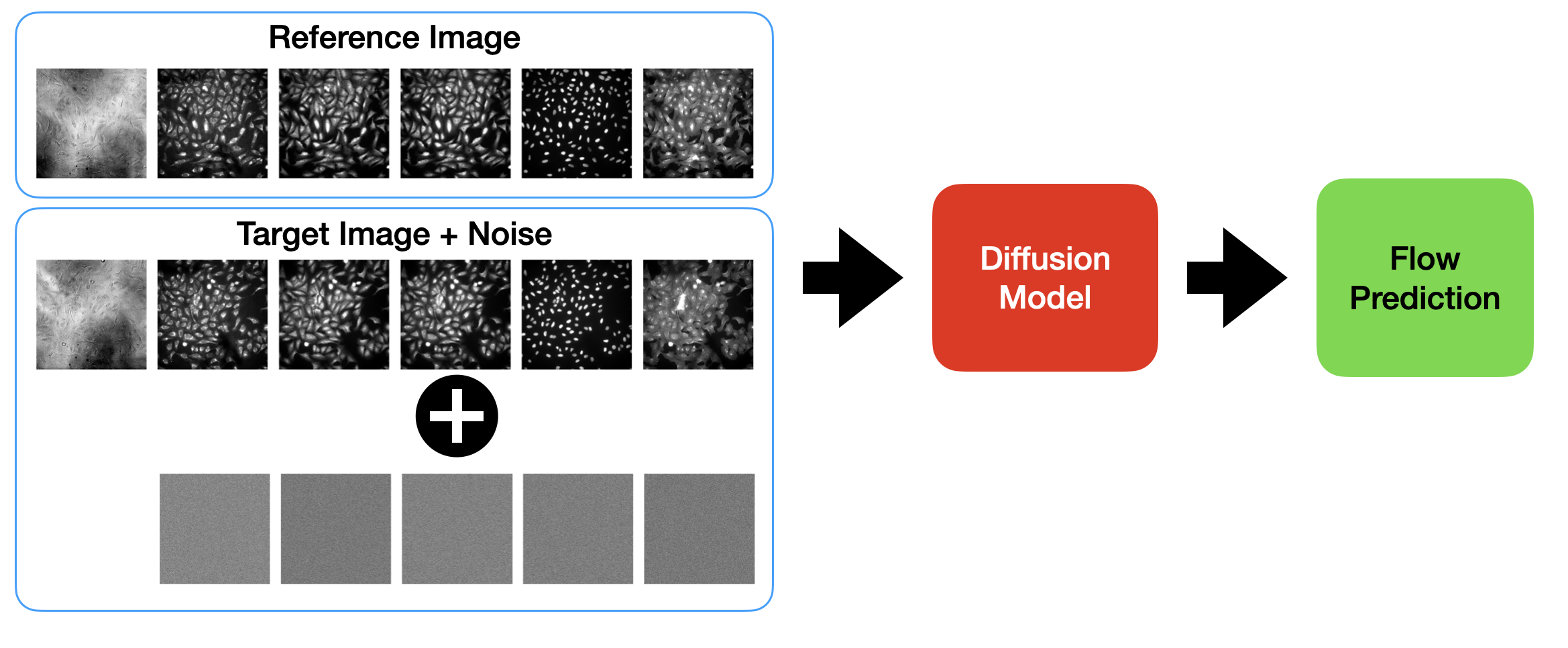}
    \caption{A visual depiction of our reference consistent architecture. At training time reference images are views sampled from the same image, at inference time for timelapse generation reference images are the first generated frame.}
    \label{fig:model}
\end{figure}
We use a UNet based diffusion model operating directly on pixels. We use the $5$ channel (DNA, RNA, ER, AGP, Mito) cell paint protocol \citep{cimini2023optimizing}.

The model input has $12$ total channels. The first 6 channels are paired (brightfield, cellpaint) channels for a image to be generated - we call these the target channels, the next $6$ channels are (brightfield, cellpaint) channels from a reference image. These can also be zeroed out to not use a reference which we do in $10$ percent of training images. 

The model has $2$ conv layers for downsampling, a final conv layer with full attention at the lowest resolution and two up-convs. This is very similar to the architecture of Stable Diffusion \citep{rombach2022high, podell2023sdxl} except in pixel space rather than the latent space of a VAE.

We use the  images from the Broad Cell Paint Gallery (\href{https://github.com/broadinstitute/cellpainting-gallery}{CPG}) \citep{weisbart2024cell}. The CPG contains hundreds of terabytes of cell painting images collected from multiple screens. The CPG subsets we use are mostly images of U2OS cells with some amount of other cancer lines (e.g. HeLA, A549). 

We use subsets of the CPG which have a brightfield channel: \textit{cpg0000} \citep{chandrasekaran2024three}, \textit{cpg0001} \citep{cimini2023optimizing},   \textit{cpg0002} \citep{tromans2023assessing}, \textit{cpg00016} \citep{chandrasekaran2023jump}, \textit{cpg0021-periscope} \citep{ramezani2025genome}, \textit{cpg0022} \citep{tegtmeyer2024high}. Some examples have the brightfield taken at multiple z-axis, at training time we sample a random z-axis. These datasets include over $8$ million images which is over $200$TB of data in their standard tiff format.

We perform minimal pre-processing on the images, for each channel and dataset we clip each individual image channel to the 1st and 99th percentiles of that channel's dataset level pixel values. For brightfield we use the 2 and 98 percentiles.\footnote{This choice was made from just taking clips and looking at the resulting images, it is likely possible to optimize this part of the pipeline further using a more principled approach.} We renormalize the pixel intensities to $[0,1]$, we take the square root \citep{anscombe1948transformation}, and then rescale to $[-1,1]$. We perform random augmentations (rotation, random zoom, etc...), and either take a $512$ square crop at random from the image or take a center crop resized to $512$ square.  

In the forward process we choose a step $t$ in the interval of $[0,1]$ discretized into $1000$ steps. The target cell paint channels are convolved with Gaussian noise $tc + (1-t)\mathcal{N}(0,1).$ 

The model receives the noised input, the target brightfield (unnoised), the reference example (if any), and the time step. During training, the reference example is another randomly augmented view taken from the same image.

\begin{figure}[h!]
    \centering
    \includegraphics[width=0.75\linewidth]{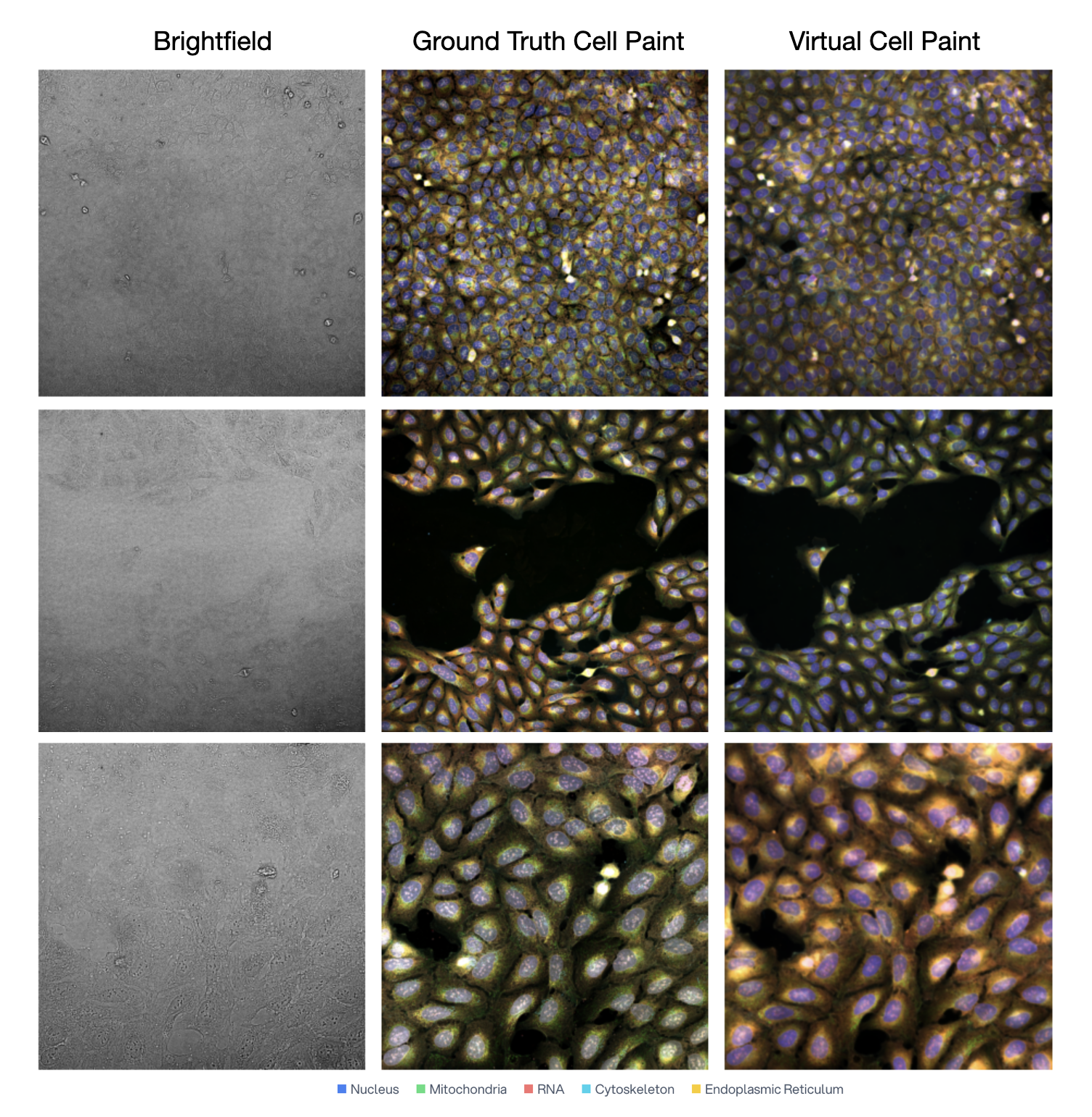}
    \caption{Examples virtual cell paint generated from MONET from brightfield image unseen during model training. There are many degrees of freedom available to a researcher doing cell painting so there is no single `correct' cell paint for a given brightfield. We use blue to render the DNA, green for mitochondria, red for RNA, teal for cytoskeleton, and yellow for the ER.} 
    \label{fig:examples}
\end{figure}

We optimize the flow matching objective \citep{lipman2022flow}, predicting the diffusion flow at each point using an MSE loss. Figure \ref{fig:model} shows our model visually.

At inference time, we perform the reverse process. For single images, we pass a fully zero conditioning. For generating timelapses from a sequence of brightfield images, we generate the first frame by looking at the brightfield of the first frame with no reference conditioning. Subsequent frames are generated by conditioning on the brightfield and generated cell paint from the first frame. We use $50$ inference steps in all generations shown here. 

\section{Single Image Generation}
We can evaluate the model first by looking at the images. Figure \ref{fig:examples} shows an example of some brightfield to virtual cellpaint generations and \url{https://www.thiscellpaintingdoesnotexist.com} shows many more. 

To get a sense of the importance of parameter scaling we trained versions of MONET ranging from 30 million to 250 million parameters using our full dataset. We discuss data scaling in a later section.

\begin{figure}[h!]
    \centering
    \includegraphics[width=0.85\linewidth]{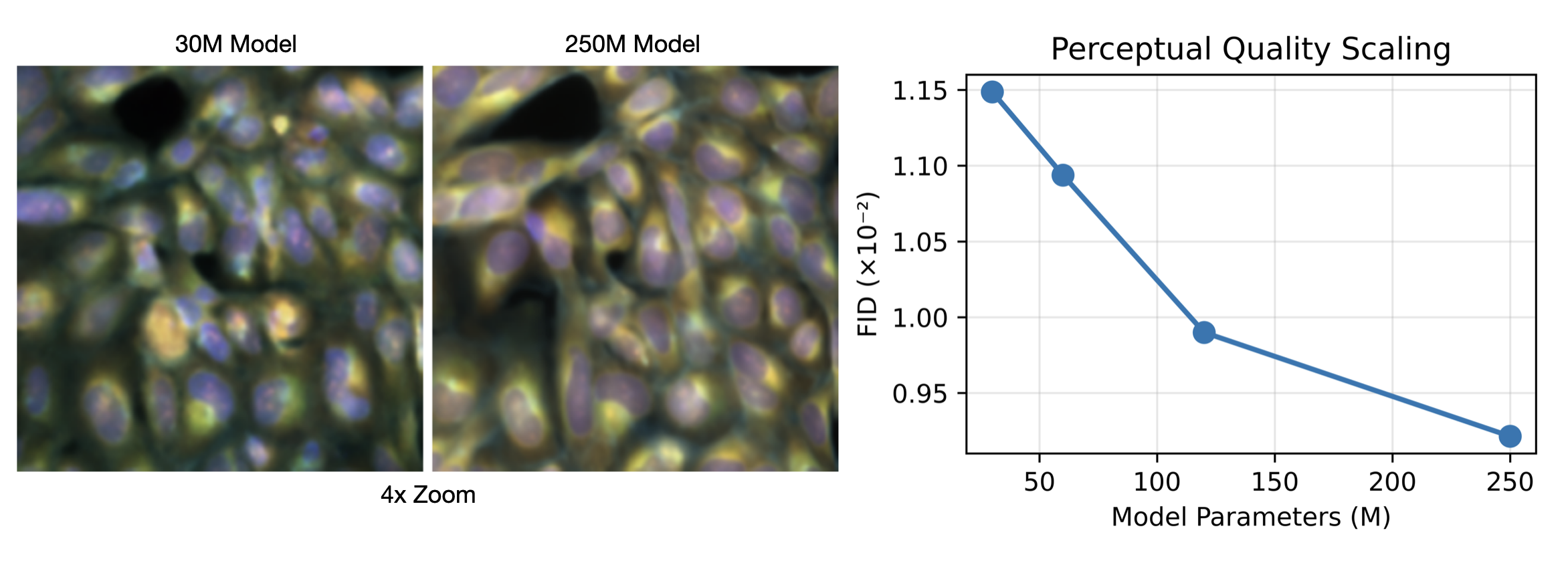}
    \caption{We trained versions of MONET ranging from 30 million to 250 million parameters. Images from smaller models are lower quality which is most visible when zoomed in. Scaling model size produces images that lead to higher scores when used to train an MOA classifier.}
    \label{fig:scaling}
\end{figure}

Visual inspection shows that images from smaller models are lower quality (e.g. blurrier), figure \ref{fig:scaling} shows a zoom in of one example image. We can quantify this by computing a perceptual distance. We use FID \citep{heusel2017gans} of RGB converted generated images compared to the ground truth. We see that the image perceptual similarity to ground truth improves as we scale models with no sign of saturating at our model scales.\footnote{We note that FID is computed from a model that is pre-trained on natural images rather than microscopy. An important future direction for the field is to develop a gold standard for perceptual metrics in the domain of biological images, however this is far outside the scope of this work.} 

However, there are many degrees of freedom available to a researcher doing cell painting - how long the dye is left, the illumination, how 5 channels are combined into 3 RGB channels for an image, etc.... Thus, there is no single `correct' cell paint for a given brightfield so metrics perceptual metrics are likely correlated with, but not perfectly representative of, model quality.

\begin{figure}[ht!]
    \centering
    \includegraphics[width=0.8\linewidth]{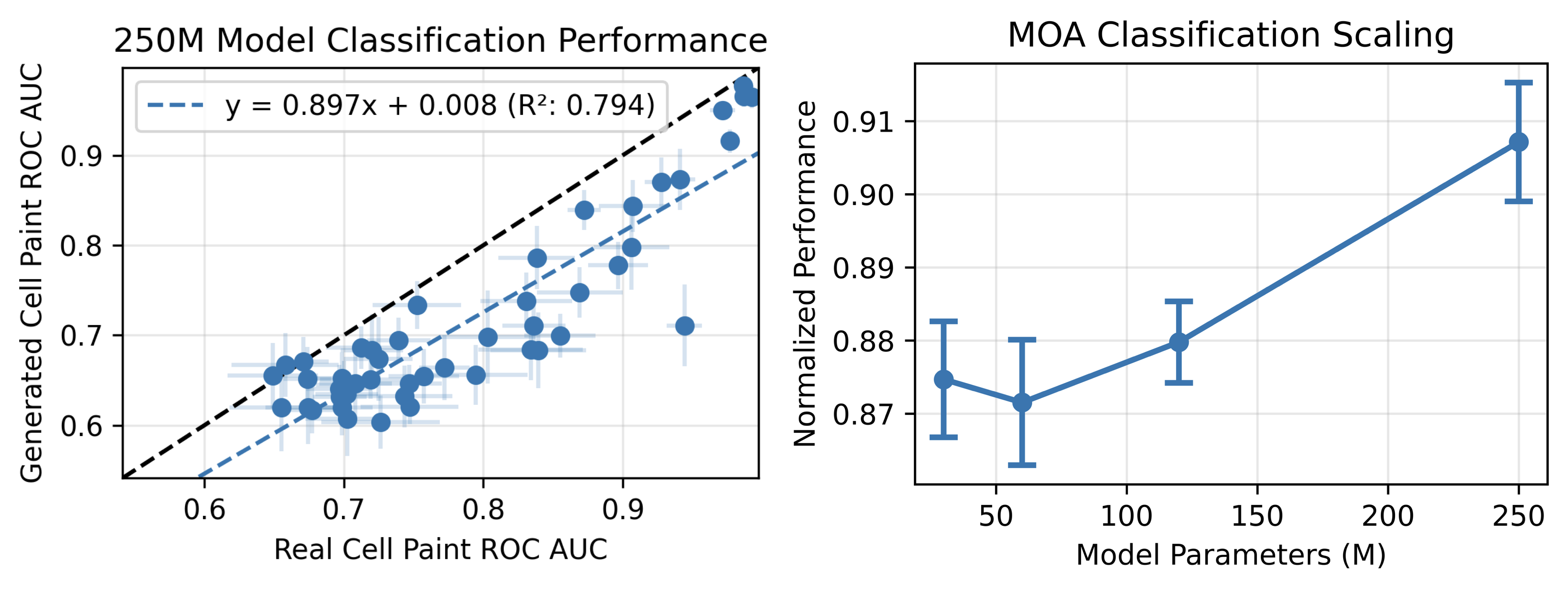}
    \caption{On the left panel each datapoint reflects the one-vs-all AUC for a single MOA. On average, MONET generations appear to capture much, but not all, information available in cell paint. Error bars reflect standard deviations in both classifiers calculated by using 10 fold cross-validation. Right panel shows how relative performance of generated cell paint scales, on average, as MONET size scales.}
    \label{fig:jumpmoa}
\end{figure}

We turn to a complementary model evaluation. We take $\sim 10,000$ images of wells that are treated with compounds from the \href{https://github.com/jump-cellpainting/JUMP-MOA}{JUMP-MOA} set using 90 compounds with 47 different mechanisms of action (MOAs). These images are not seen by the generative model during training.

We train a classifier using the base model architecture from ConvNeXT \citep{liu2022convnet} with $36$ layers and $\sim 90M$ parameters.

The classifier is trained on either the real cell paint or the virtual cell paint generated from MONET using the brightfield. We look at the one-vs-all ROC AUC per MOA of the classifiers. The performance of this classifier is a proxy for how much biologically relevant information is captured in the virtual stains and serves as a complementary metric to simply evaluating perceptual similarity.

Figure \ref{fig:jumpmoa} plots the relative performance of using real vs generated cell paints for classifier training. We see that on average the generated cell paint has approximately $90 \%$ of the AUC of the real cell paint.\footnote{We break down the model performance by MOA in more detail in Table \ref{tab:moa-results}. The heterogeneity in model predictive power across MOA is likely explained by how strongly the compound affects morpohology - for example kinesin inhibitors are easy to recognize by the classifier because they likely damage the cytoskeleton while methyltransferases modify DNA bases and have a less obvious effect on morphology.} This suggest that there are cases where virtual cell painting will provide sufficient information and there are other cases where full cell painting may be required. 

We repeat this experiment with inputs coming from MONET models trained in various sizes. As with the perceptual evaluation. We see no signs of saturation suggesting that there is more room for scaling the model and dataset. 

These results raise the question of how much the information gap between virtual and real cell paint can be narrowed by scaling or improving MONET and what information is fundamentally hidden in the brightfield that real cell painting reveals.

Interestingly, we see that a classifier trained on the generated cell paint outperforms a classifier trained only on the brightfield (Table \ref{tab:moa-results}). This suggests that virtual cell paint might form a better featurization than brightfield out of the box by highlighting subtle features automatically, this is similar to how features derived from pre-trained vision models are more efficient than simply using the pixels directly.

\section{Timelapse Generation}

To analyze the effect of our consistency architecture we used an internal dataset of U2OS cells imaged with brightfield every $10$ minutes for $24$ hours. The dataset includes $64$ wells (out of a $96$ well plate) with $9$ FOV per well. The dataset is a part of a larger experiment on cell growth, so some of the wells are treated with compounds but the details of these compounds is irrelevant for the purely visual evaluation exercise here.

\begin{figure}[ht!]
    \centering
    \includegraphics[width=0.9\linewidth]{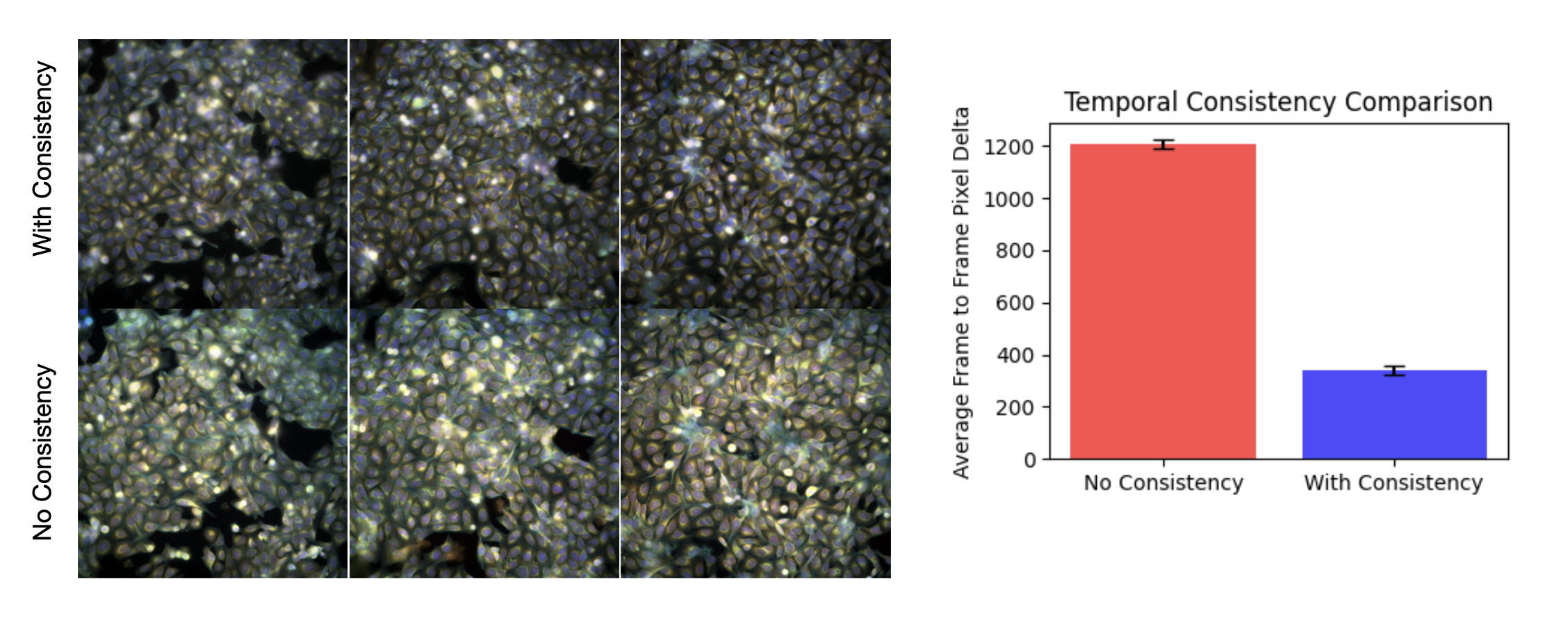}
    \caption{Generating time lapses without enforcing consistency leads to large frame-to-frame changes creating artifacts in resulting videos which can be seen in indivdual frames (left) or by looking at average frame-to-frame pixel-level deltas (right). We use blue to render the DNA, green for mitochondria, red for RNA, teal for cytoskeleton, and yellow for the ER.}
    \label{fig:consist}
\end{figure}

We show the timelapses for these cells at \url{https://www.thiscellpaintingdoesnotexist.com}.

We generated timelapses from brightfield using both our consistency architecture and by simply generating each frame independently as if they are single images. We see that independent generation leads to large amounts of flickering artifacts, we show $3$ frames from the same time lapse in Figure \ref{fig:consist}, the inconsistency artifacts are much starker in full videos such as \url{https://morelayers.ai/consistency_comparison.mp4}.

One way to quantify this pattern is to compute the MSE for two adjacent frames $(f_t, f_{t+1})$ averaged across channels generated with and without reference consistency. Figure \ref{fig:consist} shows this quantity averaged over $10$ time lapses of $\sim200$ frames each, showing that across the entire corpus indeed there is, on average, much larger frame to frame variability when generated without consistency than with consistency.

\section{Out of Sample Generalization}
\begin{figure}[h!]
    \centering
    \includegraphics[width=0.8\linewidth]{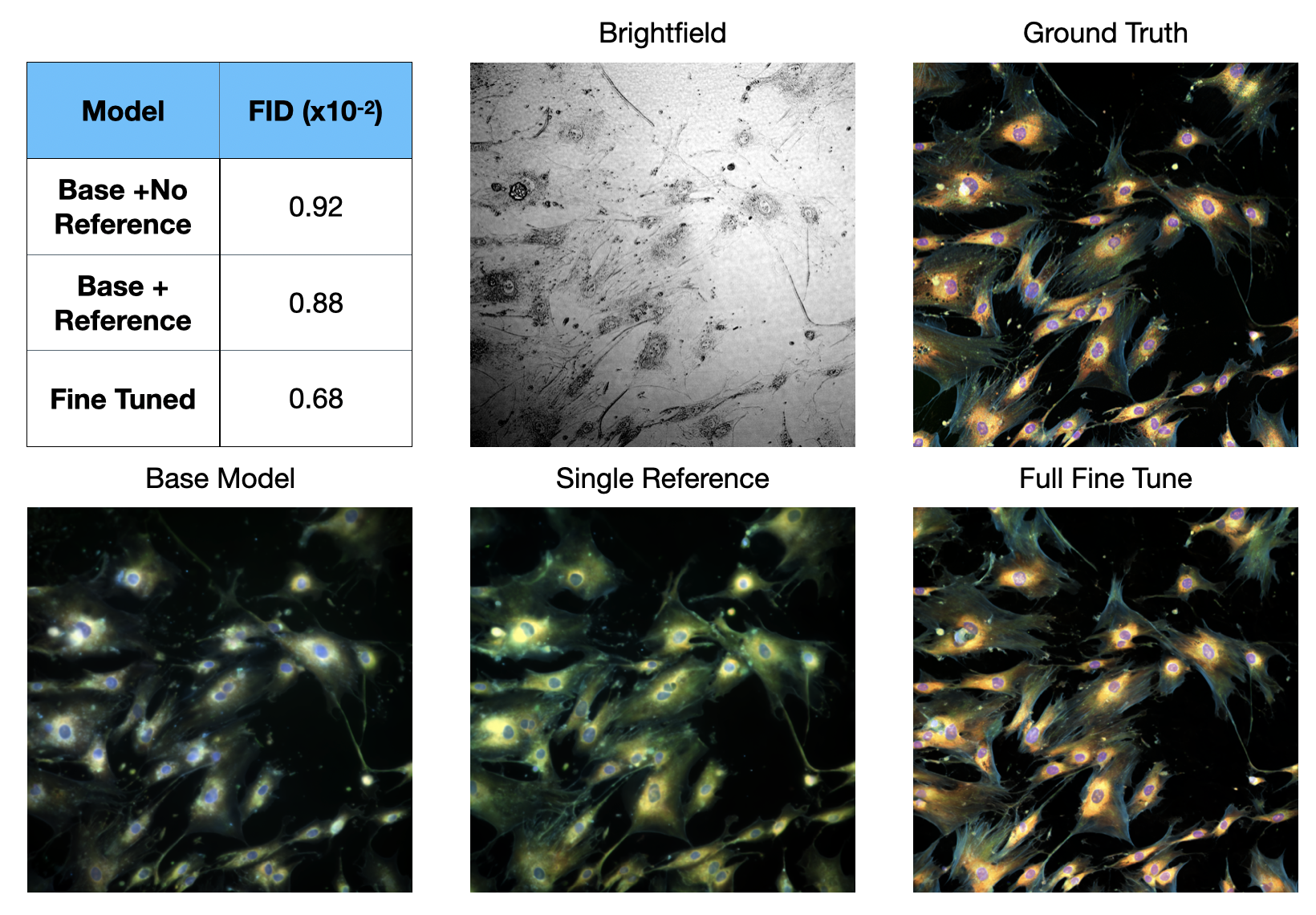}
    \caption{We collect a dataset of cell painted human fibroblasts on our imaging hardware. We see that the base MONET (trained only on cancer lines) transfers moderately. A small amount of fine tuning on the new data distribution leads to much better generation. Reference consistency allows some amount of domain adaptation via in-context learning. We use blue to render the DNA, green for mitochondria, red for RNA, teal for cytoskeleton, and yellow for the ER.}
    \label{fig:finetune}
\end{figure}

We tested out-of-the-box model generalization by using one of our proprietary datasets of cell painted human fibroblasts (a cell line the model has not seen in training). We use approximately $~7200$ images in a fine tuning training set and $\sim 800$ held out images in a test set from which we compute FID.

The CPG training data data was generated on legacy high-content imaging systems rather than the Yokogawa CQ1 spinning-disk confocal used in our pipeline. This means the pre-training data differ in optical resolution, point-spread function, and illumination uniformity. 

In Figure \ref{fig:finetune} we see that the MONET learns generalizable morphological representations, but these systematic differences in hardware, together with changes in cell type and magnification, necessitate fine-tuning on CQ1-acquired images to adapt the model to our acquisition domain fully. 

However, we also find that passing a single example from our new domain as reference gives the model some ability to adapt via a form of in-context learning. We view this latter finding as an extremely exciting direction for future research.

These results underscore the lesson that machine learning models often require domain-specific data and either some form of inference-time adaptation or explicit fine-tuning to achieve reliable performance when underlying distributions change.

\section{Conclusion}
We have presented MONET - a model to perform virtual cell painting using both still images and time lapses. MONET is not intended to be a full replacement for real staining, but a tool to augment researcher capabilities.  We have shown that the cell paintings generated by MONET contain much of the information in ground truth cell paintings.. In addition, MONET enables cell painting of time-lapses, unlocking information hidden in cell dynamics. We hope that our work helps to contribute to the computational study of cell morphology.


\newpage

\begin{table}[htbp]
\centering
\caption{Classification results broken down by MOA and model size.}
\label{tab:moa-results}
\begin{tabular}{|l|c|c|c|c|c|c|}
\hline
MOA & 30M & 60M & 120M & 250M & Brightfield & True Cell Paint \\
\hline
tumor necrosis factor production inhibitor & 0.601 & 0.619 & 0.624 & 0.655 & 0.534 & 0.649 \\
no treatment & 0.594 & 0.568 & 0.562 & 0.620 & 0.537 & 0.655 \\
histone lysine methyltransferase inhibitor & 0.649 & 0.619 & 0.620 & 0.667 & 0.537 & 0.658 \\
ubiquitin specific protease inhibitor & 0.598 & 0.627 & 0.614 & 0.671 & 0.569 & 0.671 \\
protein tyrosine kinase inhibitor & 0.644 & 0.636 & 0.631 & 0.652 & 0.522 & 0.674 \\
tricyclic antidepressant & 0.598 & 0.627 & 0.606 & 0.620 & 0.492 & 0.675 \\
androgen receptor modulator & 0.603 & 0.625 & 0.595 & 0.617 & 0.557 & 0.677 \\
phosphodiesterase inhibitor & 0.603 & 0.631 & 0.617 & 0.640 & 0.547 & 0.697 \\
CDC inhibitor & 0.596 & 0.616 & 0.626 & 0.632 & 0.569 & 0.697 \\
beta-catenin inhibitor & 0.612 & 0.599 & 0.602 & 0.620 & 0.589 & 0.699 \\
hepatocyte growth factor receptor inhibitor & 0.667 & 0.626 & 0.657 & 0.652 & 0.520 & 0.699 \\
smoothened receptor agonist & 0.628 & 0.604 & 0.660 & 0.645 & 0.570 & 0.700 \\
HMGCR inhibitor & 0.610 & 0.639 & 0.591 & 0.634 & 0.496 & 0.702 \\
JAK inhibitor & 0.597 & 0.644 & 0.623 & 0.607 & 0.485 & 0.703 \\
antihistamine & 0.663 & 0.642 & 0.630 & 0.647 & 0.532 & 0.708 \\
acetylcholine receptor antagonist & 0.678 & 0.661 & 0.674 & 0.686 & 0.624 & 0.712 \\
phospholipase inhibitor & 0.641 & 0.626 & 0.622 & 0.651 & 0.533 & 0.719 \\
LXR agonist & 0.641 & 0.628 & 0.627 & 0.684 & 0.521 & 0.720 \\
AMPK inhibitor & 0.646 & 0.628 & 0.643 & 0.674 & 0.556 & 0.725 \\
DYRK inhibitor & 0.602 & 0.566 & 0.604 & 0.604 & 0.507 & 0.726 \\
IGF-1 inhibitor & 0.626 & 0.658 & 0.653 & 0.695 & 0.545 & 0.739 \\
BCL inhibitor & 0.641 & 0.603 & 0.635 & 0.632 & 0.515 & 0.743 \\
MAP kinase inhibitor & 0.626 & 0.615 & 0.651 & 0.647 & 0.564 & 0.747 \\
MEK inhibitor & 0.631 & 0.604 & 0.653 & 0.620 & 0.527 & 0.747 \\
PARP inhibitor & 0.725 & 0.712 & 0.685 & 0.734 & 0.565 & 0.752 \\
protein arginine N-methyltransferase inhibitor & 0.615 & 0.630 & 0.596 & 0.655 & 0.565 & 0.757 \\
p38 MAPK inhibitor & 0.582 & 0.616 & 0.608 & 0.664 & 0.530 & 0.772 \\
pyruvate dehydrogenase kinase inhibitor & 0.658 & 0.634 & 0.648 & 0.656 & 0.546 & 0.795 \\
EGFR inhibitor & 0.709 & 0.699 & 0.708 & 0.698 & 0.557 & 0.803 \\
DNA inhibitor & 0.719 & 0.717 & 0.715 & 0.738 & 0.550 & 0.831 \\
bromodomain inhibitor & 0.663 & 0.655 & 0.653 & 0.684 & 0.532 & 0.834 \\
hypoxia inducible factor inhibitor & 0.692 & 0.679 & 0.694 & 0.710 & 0.538 & 0.836 \\
Bcr-Abl kinase inhibitor & 0.744 & 0.758 & 0.792 & 0.786 & 0.611 & 0.838 \\
p21 activated kinase inhibitor & 0.633 & 0.640 & 0.642 & 0.683 & 0.520 & 0.839 \\
FGFR inhibitor & 0.697 & 0.671 & 0.691 & 0.700 & 0.568 & 0.855 \\
RAF inhibitor & 0.706 & 0.707 & 0.743 & 0.748 & 0.595 & 0.869 \\
phosphoinositide dependent kinase inhibitor & 0.775 & 0.768 & 0.769 & 0.839 & 0.647 & 0.872 \\
glycogen synthase kinase inhibitor & 0.730 & 0.756 & 0.781 & 0.778 & 0.583 & 0.896 \\
inosine monophosphate dehydrogenase inhibitor & 0.768 & 0.770 & 0.782 & 0.798 & 0.558 & 0.906 \\
HDAC inhibitor & 0.762 & 0.785 & 0.799 & 0.844 & 0.623 & 0.907 \\
CDK inhibitor & 0.825 & 0.828 & 0.820 & 0.870 & 0.644 & 0.928 \\
mTOR inhibitor & 0.804 & 0.814 & 0.855 & 0.873 & 0.639 & 0.941 \\
histone lysine demethylase inhibitor & 0.681 & 0.703 & 0.668 & 0.711 & 0.548 & 0.944 \\
kinesin inhibitor & 0.939 & 0.935 & 0.937 & 0.950 & 0.667 & 0.972 \\
JNK inhibitor & 0.875 & 0.870 & 0.885 & 0.916 & 0.694 & 0.977 \\
CHK inhibitor & 0.973 & 0.970 & 0.972 & 0.977 & 0.711 & 0.986 \\
Aurora kinase inhibitor & 0.953 & 0.932 & 0.960 & 0.966 & 0.543 & 0.987 \\
MDM inhibitor & 0.961 & 0.903 & 0.953 & 0.965 & 0.575 & 0.993 \\
\hline
\end{tabular}
\end{table}

\end{document}